# Active Learning with Expert Advice

**Peilin Zhao**
Nanyang Technological University
Singapore 639798
peilinzhao@ntu.edu.sg

**Steven C.H. Hoi**
Nanyang Technological University
Singapore 639798
chhoi@ntu.edu.sg

**Jinfeng Zhuang**
Microsoft Corporation
USA 98006
jeffzh@microsoft.com

## Abstract

Conventional learning with expert advice methods assume a learner is always receiving the outcome (e.g., class labels) of every incoming training instance at the end of each trial. In real applications, acquiring the outcome from oracle can be costly or time consuming. In this paper, we address a new problem of active learning with expert advice, where the outcome of an instance is disclosed only when it is requested by the online learner. Our goal is to learn an accurate prediction model by asking the oracle the number of questions as small as possible. To address this challenge, we propose a framework of active forecasters for online active learning with expert advice, which attempts to extend two regular forecasters, i.e., Exponentially Weighted Average Forecaster and Greedy Forecaster, to tackle the task of active learning with expert advice. We prove that the proposed algorithms satisfy the Hannan consistency under some proper assumptions, and validate the efficacy of our technique by an extensive set of experiments.

## 1 Introduction

Learning with expert advice has been extensively studied for years in literature [19, 3, 13, 1]. Typically, a conventional learning with expert advice task assumes an online learner acts in an environment with a pool of experts. At each trial, the learner receives an incoming training instance, and must make a prediction on this instance based on the predictions made by the pool of experts. The outcome of the incoming instance will be disclosed by acquiring the feedback from an oracle after the learner has made the prediction, which in turn determines the incurred losses of the learner and the experts as well. The goal of this problem is to enable the online learner be able to make a prediction as accurate as possible. This framework was first introduced by Littlestone and Warmuth [19], who proposed the well-known weighted majority voting algorithms. Over the past decades, the similar problem has been extensively explored by other studies in literature, including Cesa-Bianchi et al. [3, 2], Freund and Schapire [9], Foster and Vohra [8], Haussler et al. [12], Vovk [22] etc.

The existing learning with expert advice methods assume the outcome (e.g., the true class label) of every incoming training instance will be *always* disclosed from an oracle at each trial. However, requesting the outcome of an instance from the oracle is often expensive or time consuming in many real-world applications. Unlike the conventional approaches, this paper investigates a new framework of active learning with expert advice, in which the outcome of an incoming instance may or may not be disclosed at each trial, depending on if the learner decides to make a request to the oracle. The goal of active learning with expert advice is to train the online learner to make an accurate prediction by making the number of requests to the oracle as small as possible, which is potentially applied for improving online classification with multiple kernel learning [14]. This problem is very challenging because on one hand we need to design an effective strategy to build the online learner for the training instance whenever its outcome is disclosed, and on the other hand, we must decide when the online learner should make a request for an incoming instance.

To overcome the challenge of active learning with expert advice, we present a framework of Active Forecaster algorithms and proposed two specific algorithms: (i) active weighted average forecaster and (ii) active greedy forecaster. We analyze the theoretical regret bound of the proposed algorithms, and validate their empirical efficacy via an extensive set of experiments.

The rest of this paper is organized as follows. Section 2

introduces the problem setting of learning with expert advice and the greedy forecaster algorithm. Section 3 presents the active greedy forecaster algorithm for active learning with expert advice and analyzes its theoretical performance. Section 4 shows our experimental results, and Section 5 concludes this work.

## 2 Problem Setting and Background

Specifically, we considered solving online classification problem through learning with expert advice. Online classification has been extensively studied in machine learning in the past few years [21, 4, 26, 24] for differen problems, like sentiment detection [17], cost-sensitive classification [23], feature selection [25] and etc. To solve online classification problem, the problem setting of a typical learning with expert advice task is as follows. Consider an unknown sequence of instances $\mathbf{x}_1, \ldots, \mathbf{x}_T \in \mathbb{R}^d$, a decision maker termed as "forecaster" aims to predict the outcomes (e.g., class labels) of every incoming instance $\mathbf{x}_t$. The forecaster sequentially computes its predictions based on the predictions from a set of $N$ reference forecasters called as "experts". Specifically, at the $t$-th round, after receiving an instance $\mathbf{x}_t$, the forecaster first accesses to the predictions made by the set of experts $\{f_{i,t} : \mathbb{R}^d \to [0,1] | i = 1, \ldots, N\}$, and then computes its own prediction $p_t \in [0,1]$ based on the predictions of the $N$ experts. After $p_t$ is computed, the true outcome $y_t \in \{0, 1\}$ is disclosed.

With the true outcomes revealed from the environment/oracle, the prediction performance of the forecaster and experts can be scored by some nonnegative loss function between $p_t$ and $y_t$, e.g., the absolute loss that is defined as $\ell(p_t, y_t) = |p_t - y_t|$. We can further calculate the cumulative loss experienced by each expert and the forecaster respectively as follows:

$$L_{i,t} = \sum_{j=1}^{t} \ell(f_i(\mathbf{x}_j), y_j), \quad L_t = \sum_{j=1}^{t} \ell(p_j, y_j)$$

The loss difference between the forecaster and the expert is known as the "regret", i.e.,

$$R_{i,t} = L_t - L_{i,t}, i = 1, \ldots, N \quad (1)$$

The goal of learning the forecaster is to make the regret with respect to each expert as small as possible, which is equivalent to minimizing the overall regret $R_T$, i.e.,

$$R_T = \max_{1 \le i \le N} R_{i,T} = L_T - \min_{1 \le i \le N} L_{i,T} \quad (2)$$

In general, we wish to design an ideal forecaster that can achieve a vanishing per-round regret, i.e.,

$$R_T = o(T) \iff \lim_{T \to \infty} \frac{1}{T} \left( L_T - \min_{1 \le i \le N} L_{i,T} \right) = 0$$

The above property is known as the Hannan consistency [11]. A forecaster satisfying this property is called a Hannan-consistent forecaster [11, 2].

To solve the above task of learning with expert advice, a natural framework is based on the weighted average prediction strategy. Specifically, at time $t$, the forecaster makes its own prediction as:

$$p_t = \frac{\sum_{i=1}^{N} w_{i,t-1} f_i(\mathbf{x}_t)}{\sum_{i=1}^{N} w_{i,t-1}} \quad (3)$$

where $w_{i,t-1}$ are the combination weights assigned to the experts at time t-1. The intuitive idea of learning the combination weights is to assign large weights for those experts of low regrets/loss and small weights for those of high regrets/loss.

Next we introduce a special case that leads to the well-known forecaster, called "Exponentially Weighted Average Forecaster" (EWAF). In particular, by choosing $w_{i,t-1} = \exp(\eta L_{i,t-1})/\sum_{j=1}^{N} \exp(\eta L_{j,t-1})$, where $\eta$ is a positive parameter, the EWAF strategy makes the following prediction:

$$p_t = \frac{\sum_{i=1}^{N} \exp(-\eta L_{i,t-1}) f_i(\mathbf{x}_t)}{\sum_{i=1}^{N} \exp(-\eta L_{i,t-1})}, \quad (4)$$

In addition to the weighted average forecaster, we also consider another kind of forecaster, known as the "Greedy Forecaster" (GF), which makes the following prediction:

$$p_t = \pi_{[0,1]} \left( \frac{1}{2} + \frac{1}{2\eta} \ln \frac{\sum_{i=1}^{N} \exp(-\eta L_{i,t-1} - \eta \ell(f_i(\mathbf{x}_t), 1))}{\sum_{i=1}^{N} \exp(-\eta L_{i,t-1} - \eta \ell(f_i(\mathbf{x}_t), 0))} \right),$$

where $\pi_{[0,1]}(\cdot) = \max(0, \min(1, \cdot))$. According to the existing studies [2], we have the following theorem of regret bounds of the above EWAF and GF algorithms:

**Theorem 1.** *Let the loss function $\ell(p, y) = |p - y|$. Then, for any $T$ and $\eta > 0$, and for all $y_1, \ldots, y_T \in \{0, 1\}$, the regrets of both the EWAF and GF algorithms satisfy*

$$R_T = L_T - \min_{1 \le i \le N} L_{i,T} \le \frac{\ln N}{\eta} + \frac{\eta T}{8}$$

*In particular, by choosing $\eta = \sqrt{8 \ln N / T}$, the upper bound of the regret becomes $\sqrt{(T/2) \ln N}$.*

The above theorem shows both the EWAF and GF algorithms satisfy the Hannan consistency, i.e., $R_T \le o(T)$, which guarantees that the actual per-round regret $R_T/T$ becomes negligible as $T$ grows.

## 3 Active Learning with Expert Advice

In this section, we address a new problem of active learning with expert advice. Unlike the above regular learning with expert advice task where the outcome of every incoming instance is *always* revealed to the online learner, in an active learning with expert advice task, the outcome of an incoming instance is *only* revealed whenever the learner has made a request for acquiring the label from the environment/oracle. In this section, we aim to develop a framework of *active forecasters* to tackle the challenging task of active learning with expert advice.

We first introduce binary variables $z_s \in \{0,1\}, s = 1, \ldots, t$ to indicate if an active forecaster has decided to request the class label of an incoming instance received at $s$-th trial. We denoted by $\widehat{L}_{i,t}$ the loss function experienced by the active learner w.r.t. the $i$th expert, i.e., $\widehat{L}_{i,t} = \sum_{s=1}^{t} \ell(f_i(\mathbf{x}_s), y_s) z_s$.

Hence, the class label for the $t$-th example predicted by the active forecaster, denoted by $\widehat{p}_t$, is computed as $\widehat{p}_t = \pi_{[0,1]}(\bar{p}_t)$, where $\bar{p}_t$ is computed by different approaches for different forecasters:

$$\bar{p}_t = \frac{\sum_{i=1}^{N} \exp(-\eta \widehat{L}_{i,t-1}) f_i(\mathbf{x}_t))}{\sum_{i=1}^{N} \exp(-\eta \widehat{L}_{i,t-1})} \text{ (EWAF);}$$

$$\bar{p}_t = \frac{1}{2} + \frac{1}{2\eta} \ln \frac{\sum_{i=1}^{N} \exp[\eta(-\widehat{L}_{i,t-1} - \ell(f_i(\mathbf{x}_t), 1))]}{\sum_{i=1}^{N} \exp[\eta(-\widehat{L}_{i,t-1} - \ell(f_i(\mathbf{x}_t), 0))]} \text{ (GF).}$$

In the above formula of EWAF, since $\bar{p}_t \in [0,1]$, we always have $\widehat{p}_t = \pi_{[0,1]}(\bar{p}_t) = \bar{p}_t$.

The key challenge for active learning with expert advice is to decide when the active forecaster should or should not make a request to acquire the class label w.r.t. an incoming instance from the environment/oracle. A naive solution is to consider a random sampling approach, which however may not be effective enough (this will be considered as a baseline for comparison in our empirical study). To tackle this challenge, our key motivation is to find some appropriate *confidence condition* such that it helps the online learner decide when we could skip the request of a label whenever the *confidence condition* is satisfied. To this end, we propose an idea to seek the confidence condition by estimating the difference between $p_t$ and $\widehat{p}_t$. Intuitively, the smaller the difference, the more confident we have for the prediction made by the forecaster. Before introducing our proposed confidence conditions, for convenience of presentation, we introduce a notation: $\widehat{H}_{i,t} = \sum_{s=1}^{t}(1-z_s)\ell(f_i(\mathbf{x}_s), y_s)$. It is easy to see $L_{i,t} = \widehat{L}_{i,t} + \widehat{H}_{i,t}$.

*Active Exponentially Weighted Average Forecaster (AEWAF).* We present a confidence condition for AEWAF in the following theorem, which guarantees a small difference between $p_t$ and $\widehat{p}_t$.

**Theorem 2.** *For a small constant $\delta > 0$, $\max_{1 \leq i,j \leq N} |f_i(\mathbf{x}_t) - f_j(\mathbf{x}_t)| \leq \delta$ implies $|p_t - \widehat{p}_t| \leq \delta$.*

*Proof.* For the AEWAF strategy, the distance between $p_t$ and $\widehat{p}_t$ is computed as follows:

$$|p_t - \widehat{p}_t|$$
$$= \left| \frac{\sum_{i=1}^{N} \exp(-\eta L_{i,t-1}) f_i(\mathbf{x}_t)}{\sum_{i=1}^{N} \exp(-\eta L_{i,t-1})} - \frac{\sum_{i=1}^{N} \exp(-\eta \widehat{L}_{i,t-1}) f_i(\mathbf{x}_t)}{\sum_{i=1}^{N} \exp(-\eta \widehat{L}_{i,t-1})} \right|$$
$$= \left| \frac{\sum_{i=1}^{N} \exp(-\eta \widehat{L}_{i,t-1}) \exp(-\eta \widehat{H}_{i,t-1}) f_i(\mathbf{x}_t)}{\sum_{i=1}^{N} \exp(-\eta \widehat{L}_{i,t-1}) \exp(-\eta \widehat{H}_{i,t-1})} \right.$$
$$\left. - \frac{\sum_{i=1}^{N} \exp(-\eta \widehat{L}_{i,t-1}) f_i(\mathbf{x}_t)}{\sum_{i=1}^{N} \exp(-\eta \widehat{L}_{i,t-1})} \right|$$
$$= \left| \frac{\sum_{i=1}^{N} \sum_{j=1}^{N} \gamma_{i,j,t-1}(f_i(\mathbf{x}_t) - f_j(\mathbf{x}_t))}{\sum_{i=1}^{N} \sum_{j=1}^{N} \gamma_{i,j,t-1}} \right|$$

where

$$\gamma_{i,j,t-1} = \exp(-\eta \widehat{L}_{i,t-1}) \exp(-\eta \widehat{H}_{i,t-1}) \exp(-\eta \widehat{L}_{j,t-1}).$$

Thus, if $\max_{1 \leq i,j \leq N} |f_i(\mathbf{x}_t) - f_j(\mathbf{x}_t)| \leq \delta$, it is easy to prove that $|p_t - \widehat{p}_t| \leq \delta$. □

*Active Greedy Forecaster (AGF).* We now propose a confidence condition for AGF in the theorem below, which guarantees a small difference between $\widehat{p}_t$ and $p_t$.

**Theorem 3.** *For a small constant $\delta > 0$, $\max_{1 \leq i \leq N} |f_i(\mathbf{x}_t) - \bar{p}_t| \leq \delta$ implies $|p_t - \widehat{p}_t| \leq \delta$.*

*Proof.* We can bound $p_t$ from the above as follows

$$p_t$$
$$= \pi_{[0,1]} \left( \frac{1}{2} + \frac{1}{2\eta} \ln \frac{\sum_{i=1}^{N} \exp(-\eta L_{i,t-1} - \eta \ell(f_i(\mathbf{x}_t), 1))}{\sum_{i=1}^{N} \exp(-\eta L_{i,t-1} - \eta \ell(f_i(\mathbf{x}_t), 0))} \right)$$
$$= \pi_{[0,1]} \left( \frac{1}{2} + \right.$$
$$\left. \frac{1}{2\eta} \ln \frac{\sum_{i=1}^{N} \exp(-\eta(\widehat{L}_{i,t-1} + \widehat{H}_{i,t-1}) - \eta \ell(f_i(\mathbf{x}_t), 1))}{\sum_{i=1}^{N} \exp(-\eta(\widehat{L}_{i,t-1} + \widehat{H}_{i,t-1}) - \eta \ell(f_i(\mathbf{x}_t), 0))} \right)$$
$$\leq \pi_{[0,1]} \left( \frac{1}{2} + \frac{1}{2\eta} \ln \frac{\sum_{i=1}^{N} \exp(-\eta \widehat{L}_{i,t-1} - \eta \ell(f_i(\mathbf{x}_t), 1))}{\sum_{i=1}^{N} \exp(-\eta \widehat{L}_{i,t-1} - \eta \ell(f_i(\mathbf{x}_t), 0))} \right)$$
$$+ \pi_{[0,1]} \left( \frac{1}{2\eta} \ln \frac{[\sum_{i=1}^{N} \alpha_{i,t} \exp(-\eta \widehat{H}_{i,t-1})]}{[\sum_{i=1}^{N} \beta_{i,t} \exp(-\eta \widehat{H}_{i,t-1})]} \right)$$
$$= \widehat{p}_t + \pi_{[0,1]} \left( \frac{1}{2\eta} \ln \frac{[\sum_{i=1}^{N} \alpha_{i,t} \exp(-\eta \widehat{H}_{i,t-1})]}{[\sum_{i=1}^{N} \beta_{i,t} \exp(-\eta \widehat{H}_{i,t-1})]} \right)$$

where

$$\alpha_{i,t} = \frac{\exp\left(-\eta \left[\widehat{L}_{i,t-1} + \ell(f_i(\mathbf{x}_t), 1)\right]\right)}{\sum_{j=1}^{N} \exp\left(-\eta \left[\widehat{L}_{j,n-1} + \ell(f_j(\mathbf{x}_t), 1)\right]\right)},$$

$$\beta_{i,t} = \frac{\exp\left(-\eta \left[\widehat{L}_{i,t-1} + \ell(f_i(\mathbf{x}_t), 0)\right]\right)}{\sum_{j=1}^{N} \exp\left(-\eta \left[\widehat{L}_{j,n-1} + \ell(f_j(\mathbf{x}_t), 0)\right]\right)}, \; i \in [N].$$

Since $\forall i \in [N]$

$$\frac{\alpha_{i,t}}{\beta_{i,t}} = \frac{\sum_{j=1}^{N} \exp\left(-\eta\left[\widehat{L}_{j,t-1} + \ell(f_j(\mathbf{x}_t), 0)\right]\right)}{\sum_{j=1}^{N} \exp\left(-\eta\left[\widehat{L}_{j,t-1} + \ell(f_j(\mathbf{x}_t), 1)\right]\right)} \times$$

$$\frac{\exp\left(-\eta\left[\widehat{L}_{i,t-1} + \ell(f_i(\mathbf{x}_t), 1)\right]\right)}{\exp\left(-\eta\left[\widehat{L}_{i,t-1} + \ell(f_i(\mathbf{x}_t), 0)\right]\right)}$$

$$= \frac{\exp\left(-\eta\left[\ell(f_i(\mathbf{x}_t), 1) - \ell(f_i(\mathbf{x}_t), 0)\right]\right)}{\exp\left(\eta\left(2\bar{p}_t - 1\right)\right)}$$

$$= \frac{\exp\left(\eta\left(2f_i(\mathbf{x}_t) - 1\right)\right)}{\exp\left(\eta\left(2\bar{p}_t - 1\right)\right)}$$

$$= \exp\left(2\eta\left(f_i(\mathbf{x}_t) - \bar{p}_t\right)\right) \leq \exp(2\eta\delta),$$

and $\ln x$ is an increasing function, we have

$$\ln \frac{\sum_{i=1}^{N} \alpha_{i,t} \exp(-\eta \widehat{H}_{i,t-1})}{\sum_{j=1}^{N} \beta_{j,t-1} \exp(-\eta \widehat{H}_{j,t-1})} \leq 2\eta\delta$$

and $p_t \leq \widehat{p}_t + \delta$. Similar to the above analysis, we have $p_t$ lower bounded as $p_t \geq \widehat{p}_t - \delta$. □

Based on the above analysis of the confidence conditions, we can now present the general framework of active forecasters for online active learning with expert advice, which is summarized in Algorithm 1.

---

**Algorithm 1** A Framework of Active Forecaster

**Input**: a pool of experts $f_i$, $i = 1, \ldots, N$.
**Initialize** tolerance threshold $\delta$ and $\widehat{L}_{i,t} = 0$, $i \in [N]$.
**for** $t = 1, \ldots, T$ **do**
  receive $\mathbf{x}_t$ and compute $f_i(\mathbf{x}_t)$, $i \in [N]$;
  compute $\bar{p}_t$ according to equation (5) and set $\widehat{p}_t = \pi_{[0,1]}(\bar{p}_t)$;
  **if** the *confidence condition* is satisfied **then**
    skip the label request for instance $\mathbf{x}_t$
  **else**
    request label $y_t$ and update $\widehat{L}_{i,t} = \widehat{L}_{i,t-1} + \ell(f_i(\mathbf{x}_t), y_t)$, $i \in [N]$;
  **end if**
**end for**

---

As shown in Algorithm 1, at each round, after receiving an input instance $\mathbf{x}_t$, we compute the prediction by each expert in the pool, i.e., $f_i(\mathbf{x}_t)$. Then, we examine if the confidence condition is satisfied. If so, we will skip the label request for this instance; otherwise, the learner will request the class label for this instance from the environment.

We now present a theorem about the upper bound of the regret of the two active forecasters, i.e.,

$$\widehat{R}_T = \widehat{L}_T - \min_{i \in [N]} L_{i,T}$$

where $\widehat{L}_T = \sum_{t=1}^{T} \ell(\widehat{p}_t, y_t)$, which is the overall loss experienced by the active forecaster.

**Theorem 4.** *Let the loss function $\ell(p, y) = |p - y|$, and denote by $Q$ the total number of requested labels, i.e., $Q = \sum_{t=1}^{T} z_t$, then, for any $T$, $\eta > 0$ and for all $y_1, \ldots, y_T \in \{0, 1\}$, the regret $\widehat{R}_T$ of the two Active Forecasters (AEWAF and AGF) can be bounded as:*

$$\widehat{R}_T = \widehat{L}_T - \min_{i \in [N]} L_{i,T} \leq \frac{\ln N}{\eta} + \frac{\eta T}{8} + \delta(T - Q).$$

*Proof.* Firstly according to Theorem 2 and 3, we have the following bound for $\widehat{L}_T - L_T$:

$$\begin{aligned}\widehat{L}_T - L_T &= \sum_{t=1}^{T} (\ell(\widehat{p}_t, y_t) - \ell(p_t, y_t)) \\ &\leq (T - Q) * |\widehat{p}_t - p_t| \leq \delta(T - Q).\end{aligned}$$

Combining the above result with Theorem 1, we have the regret of the Active Forecasters bounded:

$$\widehat{R}_T = \widehat{L}_T - \min_{1 \leq i \leq N} L_{i,T} = (\widehat{L}_T - L_T) + (L_T - \min_{1 \leq i \leq N} L_{i,T})$$

$$\leq \delta(T - Q) + \frac{\ln N}{\eta} + \frac{\eta T}{8}.$$

□

*Remark.* For the above theorem, if a learner requests the labels for every instance, i.e., Q=T, the bound reduces the bound of the regular forecasters. Based on the above theorem, we have the following corollary that shows the proposed Active Forecasters satisfy the Hannan consistency.

**Corollary 5.** *Consider $0 < a << T$, if we set $\eta = \sqrt{\frac{8 \ln N}{T(1+8a) - 8aQ}}$ and $\delta = a\eta$, then we have the regret achieved by the proposed algorithms bounded as $o(T)$.*

*Proof.* Following the result of Theorem 4, we have

$$\widehat{R}_T \leq \frac{\ln N}{\eta} + \frac{\eta T}{8} + \delta(T - Q)$$

$$= \frac{\ln N}{\eta} + \eta \left((a + \frac{1}{8})T - Qa\right)$$

$$= 2\sqrt{\ln N} \sqrt{T\left(a + \frac{1}{8}\right) - Qa}$$

where the last equation holds under the condition $\frac{\ln N}{\eta} = \eta\left((a + \frac{1}{8})T - Qa\right)$, i.e., $\eta = \sqrt{\frac{8 \ln N}{T(1+8a) - 8aQ}}$, and as a result $\delta = a\sqrt{\frac{8 \ln N}{T(1+8a) - 8aQ}}$. Therefore, we have $\widehat{R}_T = o(T)$. □

## 4 Experimental Results

In this section, we evaluate the empirical performance of the proposed Active Forecasters for online active learning with expert advice tasks.

### 4.1 Experts and Compared Algorithms

To construct experts for an online sequential prediction task, we choose to build the pool of experts by adopting five well-known online learning algorithms [7, 5, 20], which include: (implemented as in [16])

- PERCEPTRON: the classical Perceptron algorithm [21];

- ROMMA: the Relaxed Online Maximum Margin Algorithm [18];

- ALMA$_p(\alpha)$: the Approximate Maximal Margin Algorithm [10];

- PA: the Passive-Aggressive online learning algorithm [4];

- AROW: the Adaptive Regularization Of Weights algorithm [6].

We compare the two proposed active learning algorithms (AEWAF and AGF) with the two regular forecasters (EWAF and GF) algorithm and their random variants as well, which are listed below:

- EWAF: the Exponentially Weighted Forecaster [2];

- GF: the Greedy Forecaster algorithm [2];

- REWAF: the Random Exponentially Weighted Forecaster, a variant of EWAF, which will randomly select the indices according to uniform distribution;

- RGF: the Random Greedy Forecaster algorithm, a variant of GF, which will randomly select the indices according to uniform distribution;

- AEWAF: the proposed Active Exponentially Weighted Forecaster algorithm;

- AGF: the proposed Active Greedy Forecaster algorithm.

### 4.2 Experimental Testbed and Setup

To evaluate the performance, we conduct experiments on a variety of benchmark datasets from web machine learning repositories. Table 1 shows the details of 9 datasets used in our experiments. All of them can be downloaded from LIBSVM website [1] and UCI machine learning repository [2]. These datasets are chosen fairly randomly in order to cover various aspects of datasets.

Table 1: Datasets used in the experiments.

| Dataset | Name | # instances | # features |
|---------|------|-------------|------------|
| D1 | a8a | 32561 | 123 |
| D2 | codrna | 271617 | 8 |
| D3 | covtype | 581012 | 54 |
| D4 | gisette | 7000 | 5000 |
| D5 | magic04 | 19020 | 10 |
| D6 | mushrooms | 8124 | 112 |
| D7 | spambase | 4601 | 57 |
| D8 | svmguide1 | 7089 | 4 |
| D9 | w8a | 64700 | 300 |

All the expert algorithms learn a linear classifier for a binary classification task. The parameter $p$ and $\alpha$ in ALMA$_p(\alpha)$ are set to be 2 and 0.9 respectively. The parameter $C$ in PA is set to 5, and the parameter $\gamma$ is set to 1 for AROW. To make fair comparisons, all the compared forecasters adopt the same setup. The learning rate $\eta$ is set to $\sqrt{8 \ln N/T}$, for all the datasets and forecasters. The sampling ratio for requesting labeled data by the two random algorithms (REWAF and RGF) are set according to the ratio of labeled data requested by AEWAF and AGF using different $\delta$ values, respectively.

Each dataset is randomly divided into two subsets: a training set consisting of 20% of the entire data for training the experts; and a test set consisting of the remaining data for learning the forecasters. The five experts algorithms are applied on the training set to learn the five expert functions $\mathbf{u}_i \in \mathbb{R}^d$, $i \in [5]$, where $d$ is the dimension of the instance. To satisfy the assumptions, we adopt $f_i(\mathbf{x}) = \pi_{[0,1]}(\mathbf{u}_i^\top \mathbf{x} + 0.5)$ as the expert functions. Then we test the forecasters on the test set. All the test experiments were conducted over 20 runs of different random permutations for each test set. All the results were reported by averaging over these 20 runs. For performance metric, we evaluate the forecasters by measuring the regret rate, ratio of requested labeled data, and the running time cost.

### 4.3 Evaluation of Regular Forecasters

Table 2 summarizes the average performance of the EWAF and GF algorithms for conventional online learning with expert advice on the benchmark datasets.

---

[1] http://www.csie.ntu.edu.tw/~cjlin/libsvmtools/
[2] http://www.ics.uci.edu/~mlearn/MLRepository.html

Table 2: Evaluation of two regular forecasters (EWAF and GF) on all the datasets.

| Dataset | Alg. | Measures | |
|---|---|---|---|
| | | Regret (%) | Time (s) |
| D1 | EWAF | 0.286 ± 0.001 | 0.694 |
| | GF | 0.286 ± 0.001 | 0.841 |
| D2 | EWAF | 0.207 ± 0.001 | 1.019 |
| | GF | 0.207 ± 0.001 | 1.282 |
| D3 | EWAF | 0.066 ± 0.001 | 10.942 |
| | GF | 0.066 ± 0.001 | 13.428 |
| D4 | EWAF | 0.609 ± 0.001 | 0.574 |
| | GF | 0.609 ± 0.001 | 0.593 |
| D5 | EWAF | 0.373 ± 0.001 | 0.332 |
| | GF | 0.373 ± 0.001 | 0.414 |
| D6 | EWAF | 0.483 ± 0.001 | 0.157 |
| | GF | 0.483 ± 0.001 | 0.194 |
| D7 | EWAF | 0.758 ± 0.001 | 0.084 |
| | GF | 0.758 ± 0.001 | 0.103 |
| D8 | EWAF | 0.506 ± 0.001 | 0.123 |
| | GF | 0.506 ± 0.001 | 0.154 |
| D9 | EWAF | 0.164 ± 0.001 | 1.760 |
| | GF | 0.164 ± 0.001 | 2.031 |

Table 3: Evaluation of REWAF and AEWAF on all the dataset. R. denotes REWAF and A. denotes AEWAF. $\delta$ is set as 0.2.

| Data | Alg. | Measures | | |
|---|---|---|---|---|
| | | Regret (%) | Query (%) | Time (s) |
| D1 | R. | 0.364 ± 0.017 | 77.35 ± 0.23 | 0.685 |
| | A. | 0.292 ± 0.001 | 77.44 ± 0.01 | 0.762 |
| D2 | R. | 0.491 ± 0.025 | 42.48 ± 0.16 | 0.796 |
| | A. | 0.202 ± 0.001 | 42.46 ± 0.01 | 0.941 |
| D3 | R. | 0.088 ± 0.002 | 74.26 ± 0.05 | 10.465 |
| | A. | 0.064 ± 0.000 | 74.25 ± 0.01 | 11.958 |
| D4 | R. | 1.342 ± 0.102 | 45.40 ± 0.59 | 0.566 |
| | A. | 0.598 ± 0.003 | 45.40 ± 0.01 | 0.568 |
| D5 | R. | 0.907 ± 0.076 | 40.94 ± 0.33 | 0.256 |
| | A. | 0.344 ± 0.004 | 41.00 ± 0.01 | 0.300 |
| D6 | R. | 1.575 ± 0.067 | 10.80 ± 0.38 | 0.104 |
| | A. | 0.530 ± 0.004 | 10.63 ± 0.01 | 0.119 |
| D7 | R. | 1.062 ± 0.066 | 71.64 ± 0.73 | 0.080 |
| | A. | 0.756 ± 0.004 | 71.67 ± 0.01 | 0.091 |
| D8 | R. | 0.718 ± 0.047 | 49.36 ± 0.50 | 0.100 |
| | A. | 0.535 ± 0.006 | 49.41 ± 0.01 | 0.117 |
| D9 | R. | 0.248 ± 0.006 | 40.79 ± 0.17 | 1.489 |
| | A. | 0.169 ± 0.000 | 40.82 ± 0.01 | 1.636 |

From the experimental results in Table 2, we can draw several observations. First, the regret rates of EWAF and GF on every dataset are almost the same, which is consistent to the theoretical result that shows that they share the same regret bound. Second, EWAF consumes slightly less time cost than GF for all the cases due to the difference of their solutions. Finally, we found that the larger the dataset size, the smaller the average regret value achieved by the two algorithms. This is consistent with the Hannan property satisfied by the two algorithms, i.e., the average regret is negligible when $T$ is very large.

### 4.4 Evaluation of Active Forecasters on Fixed Ratio of Queries

In this subsection, the tolerance threshold $\delta$ for AEWAF and AGF is set as 0.2. Table 3 summarizes the average performance of the REWAF and AEWAF algorithms over the experimental datasets. From the experimental results, we can draw several observations as follows.

First of all, since we choose the sampling threshold $\rho$ according to the ratios of required labels by AEWAF using a fixed tolerances $\delta$, the differences between the ratios of requested labeled data for AEWAF and REWAF are not statistically significant, which has been verified by statistical t-tests. This implies that the statistical differences between the regret rates achieved by AEWAF and REWAF, if any, are not caused by the differences between their ratios of the requested labeled data.

Second, compared with REWAF, AEWAF achieves statistically lower regret rates on all the datasets, which validates the effectiveness of the proposed active learning strategy and also indicates the importance of exploiting the degree of agreements between different experts. In addition, AEWAF can achieve comparable regret rates with EWAF by requesting a significantly less amount of labels; while REWAF suffers significantly more regret rates by requesting the same amount of labels. This shows that AEWAF could be an attractive alternative to the EWAF in order to save the expensive labeling efforts in a real application.

Third, the time cost of the AEWAF algorithm is in general comparable to or slightly higher than that of the REWAF algorithm because the proposed confidence conditions can be evaluated rather efficiently.

Finally, we would also like to examine if the proposed active learning strategy can be generalized to different types of forecasting algorithms. To this purpose, we also evaluate the performance of the RGF and AGF algorithms. Table 4 summarizes the experimental results on all the datasets. As compared to the last experiment, similar observations can be drawn from the experimental results. We found that AEWAF and AGF request almost the same ratios of labels and achieve comparable regret rates on all the datasets; while REWAF and RGF achieve comparable regret rates, which are significantly higher than those of the two proposed active algorithms. These results indicate that the proposed active learning strategy can be generalized to different types of forecasting algorithms, and again validate the efficacy of the proposed active learning algorithms.

Table 4: Evaluation of RGF and AGF on all the dataset. R. denotes RGF and A. denotes AGF. $\delta$ is set as 0.2.

| Data | Alg. | Measures | | |
|---|---|---|---|---|
| | | Regret (%) | Query (%) | Time (s) |
| D1 | R. | 0.362 ± 0.022 | 76.61 ± 0.26 | 0.905 |
| | A. | 0.288 ± 0.003 | 76.56 ± 0.04 | 0.974 |
| D2 | R. | 0.474 ± 0.028 | 42.38 ± 0.21 | 1.363 |
| | A. | 0.199 ± 0.002 | 42.40 ± 0.01 | 1.542 |
| D3 | R. | 0.089 ± 0.002 | 74.21 ± 0.06 | 14.416 |
| | A. | 0.063 ± 0.001 | 74.21 ± 0.01 | 15.959 |
| D4 | R. | 1.344 ± 0.088 | 45.53 ± 0.71 | 0.610 |
| | A. | 0.592 ± 0.005 | 45.27 ± 0.04 | 0.611 |
| D5 | R. | 0.923 ± 0.096 | 40.13 ± 0.39 | 0.442 |
| | A. | 0.335 ± 0.009 | 40.32 ± 0.07 | 0.498 |
| D6 | R. | 1.633 ± 0.053 | 10.21 ± 0.41 | 0.206 |
| | A. | 0.543 ± 0.005 | 10.22 ± 0.07 | 0.231 |
| D7 | R. | 1.088 ± 0.080 | 71.18 ± 0.82 | 0.113 |
| | A. | 0.752 ± 0.007 | 71.27 ± 0.06 | 0.126 |
| D8 | R. | 0.726 ± 0.043 | 46.97 ± 0.48 | 0.164 |
| | A. | 0.559 ± 0.015 | 47.01 ± 0.40 | 0.185 |
| D9 | R. | 0.247 ± 0.005 | 40.55 ± 0.21 | 2.110 |
| | A. | 0.171 ± 0.001 | 40.59 ± 0.01 | 2.303 |

### 4.5 Evaluation of Active Forecasters on Varied Ratios of Queries

Firstly, Figure (1) shows the performance of the REWAF and AEWAF algorithms on mushrooms with varied ratios of queries. AEWAF outperforms REWAF with all the ratios of queries, which verifies the proposed active strategies are effective and promising.

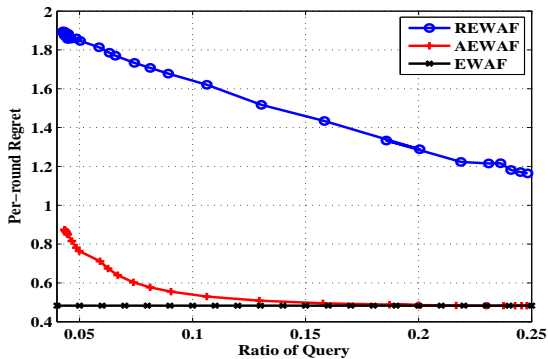

Figure 1: Comparison of REWAF and AEWAF on mushrooms.

Secondly, Figure (2) shows the performance of the RGF and AGF algorithms on mushrooms with varied ratios of queries. AEWAF outperforms REWAF with all the ratios of queries, which again verifies the proposed active strategies are effective and promising.

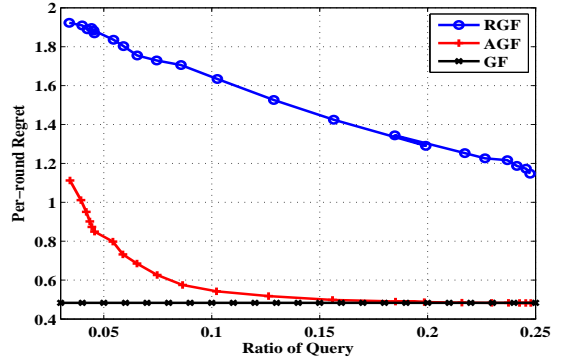

Figure 2: Comparison of RGF and AGF on mushrooms.

Finally, Figure (3) and (4) shows the comparisons of REWAF and AEWAF, RGF and AGF on all the remaining datasets, respectively, where similar observations can be found from the results.

## 5 Conclusion

This paper addressed a new problem of active learning with expert advice for online sequential prediction tasks. We proposed two novel strategies for active learning with expert advice by extending two existing forecasting algorithms in an online active learning setting. We analyze the theoretical regret bounds for the proposed algorithms, which guarantee the proposed algorithms satisfy the important Hannan consistency. We have conducted an extensive set of experiments to evaluate the efficacy of the algorithms. Promising empirical results validate the effectiveness of our technique.

Despite the encouraging results, some limitations and open challenges of the current work remain. One issue is about the settings of the learning rate $\eta$ and tolerance parameter $\delta$, which were fixed manually in our experiments. It would be more attractive if one is able to design a self-tuned strategy for the active learning task. Further, the current regret bounds may be further improved, e.g., by adopting different loss functions or other strategies. Another future work may be exploring the principles of semi-supervised learning for improving active learning with expert advice [15]. These issues can be further explored in the future work.

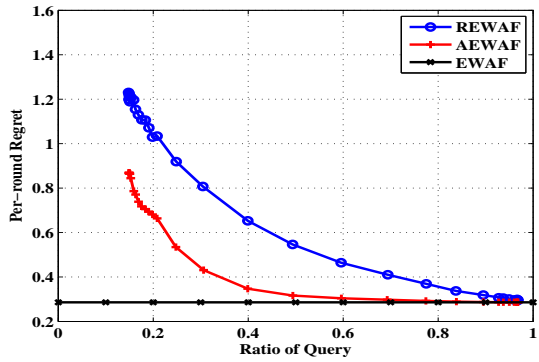
(a) a8a

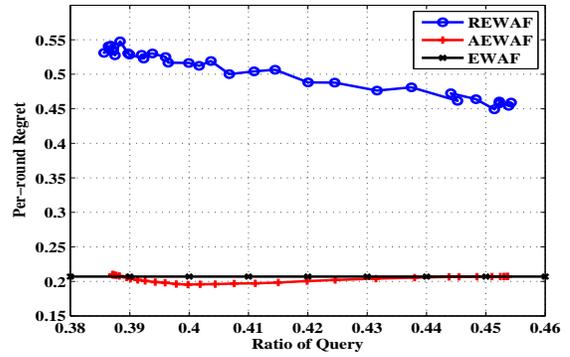
(b) codrna

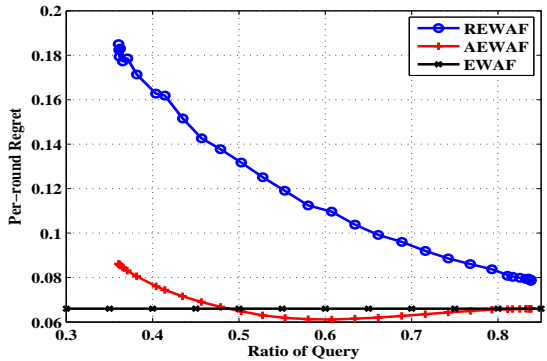
(c) covtype

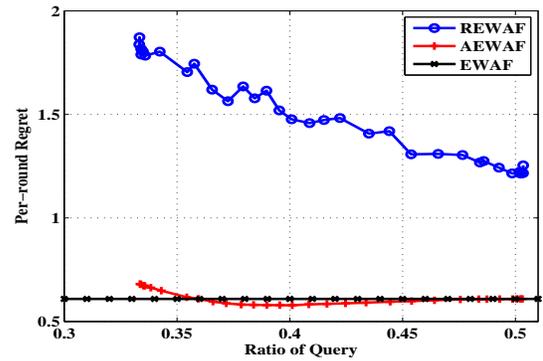
(d) gisette

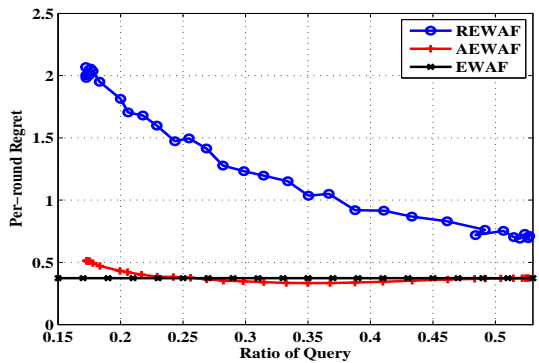
(e) magic04

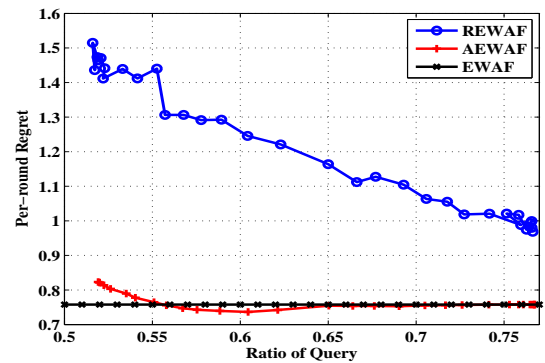
(f) spambase

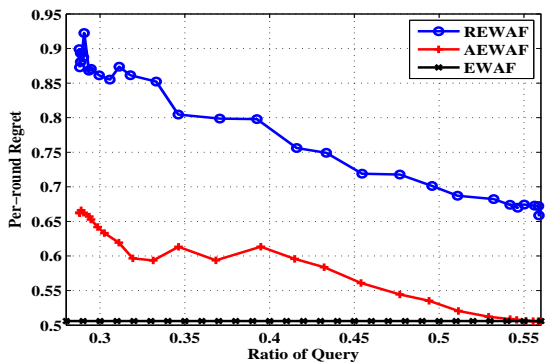
(g) svmguide1

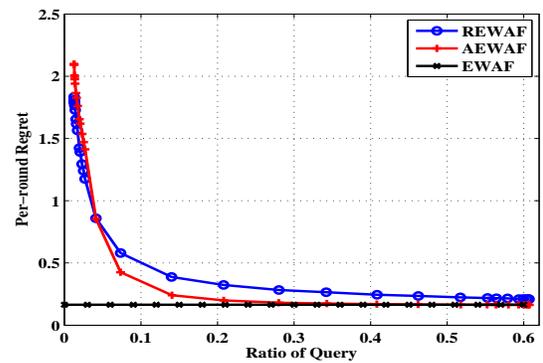
(h) w8a

Figure 3: Comparison of regret rates with respect to varied ratios of queries.

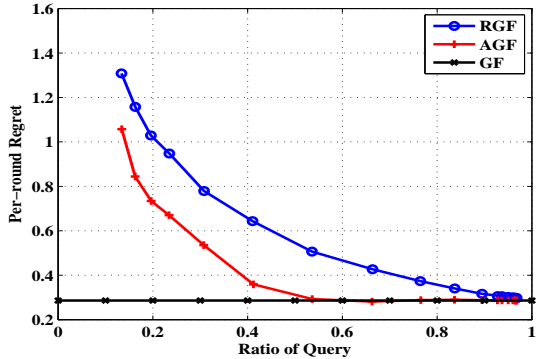

(a) a8a

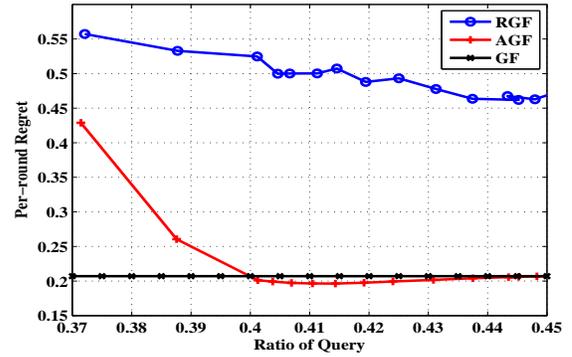

(b) codrna

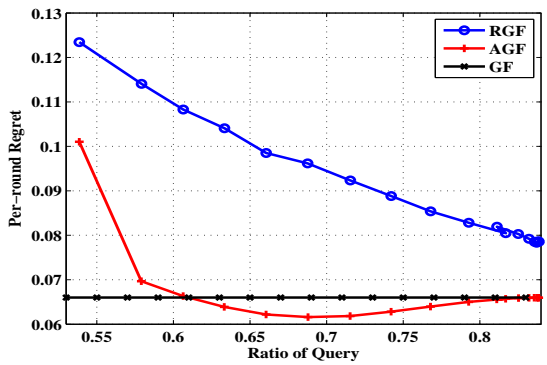

(c) covtype

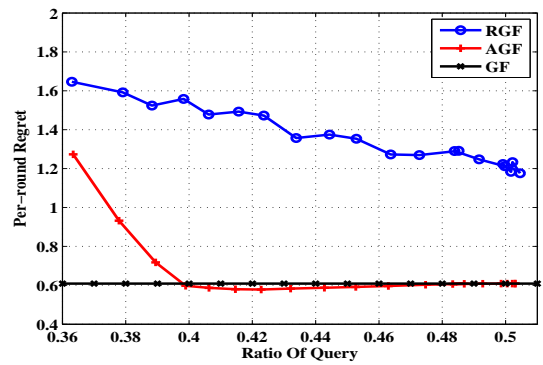

(d) gisette

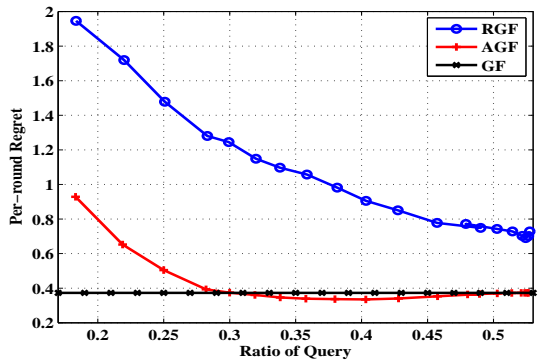

(e) magic04

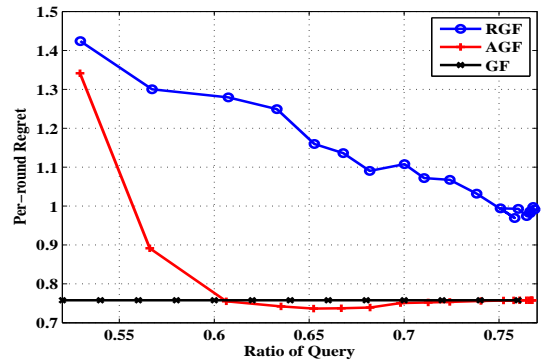

(f) spambase

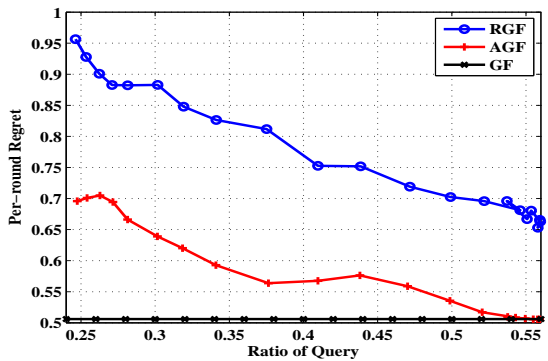

(g) svmguide1

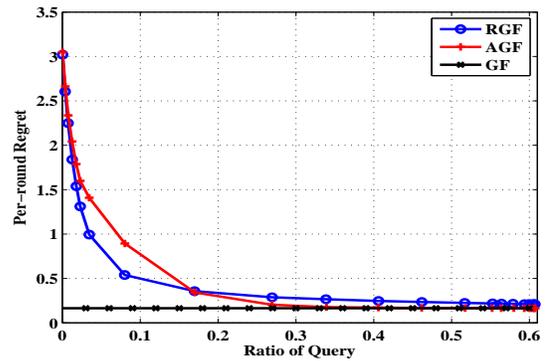

(h) w8a

Figure 4: Comparison of regret rates with respect to varied ratios of queries.

# References

[1] Olivier Bousquet and Manfred K. Warmuth. Tracking a small set of experts by mixing past posteriors. *Journal of Machine Learning Research*, 3:363–396, 2002.

[2] N. Cesa-Bianchi and G. Lugosi. *Prediction, Learning, and Games*. Cambridge University Press, 2006.

[3] Nicolò Cesa-Bianchi, Yoav Freund, David Haussler, David P. Helmbold, Robert E. Schapire, and Manfred K. Warmuth. How to use expert advice. *J. ACM*, 44(3):427–485, 1997.

[4] Koby Crammer, Ofer Dekel, Joseph Keshet, Shai Shalev-Shwartz, and Yoram Singer. Online passive-aggressive algorithms. *Journal of Machine Learning Research*, 7:551–585, 2006.

[5] Koby Crammer, Mark Dredze, and Fernando Pereira. Exact convex confidence-weighted learning. In *Advances in Neural Information Processing Systems (NIPS)*, pages 345–352, 2008.

[6] Koby Crammer, Alex Kulesza, and Mark Dredze. Adaptive regularization of weight vectors. In *NIPS*, pages 414–422, 2009.

[7] Mark Dredze, Koby Crammer, and Fernando Pereira. Confidence-weighted linear classification. In *Proceedings of the 25th International Conference on Machine Learning (ICML2008)*, pages 264–271, 2008.

[8] Dean P. Foster and Rakesh V. Vohra. A randomization rule for selecting forecasts. *Oper. Res.*, 41:704–709, July 1993.

[9] Yoav Freund and Robert E. Schapire. A decision-theoretic generalization of on-line learning and an application to boosting. *J. Comput. Syst. Sci.*, 55:119–139, August 1997.

[10] Claudio Gentile. A new approximate maximal margin classification algorithm. *Journal of Machine Learning Research*, 2:213–242, 2001.

[11] J. Hannan. Approximation to bayes risk in repeated plays. *Contributions to the Theory of Games*, 3:97–139, 1957.

[12] David Haussler, Jyrki Kivinen, and Manfred K. Warmuth. Tight worst-case loss bounds for predicting with expert advice. In *EuroCOLT*, pages 69–83, 1995.

[13] Mark Herbster and Manfred K. Warmuth. Tracking the best expert. *Machine Learning*, 32(2):151–178, 1998.

[14] Steven C. H. Hoi, Rong Jin, Peilin Zhao, and Tianbao Yang. Online multiple kernel classification. *Machine Learning*, 90(2):289–316, 2013.

[15] Steven C.H. Hoi, Rong Jin, Jianke Zhu, and Michael R Lyu. Semisupervised svm batch mode active learning with applications to image retrieval. *ACM Transactions on Information Systems (TOIS)*, 27(3):16, 2009.

[16] Steven C.H. Hoi, Jialei Wang, and Peilin Zhao. *LIBOL: A Library for Online Learning Algorithms*. Nanyang Technological University, 2012.

[17] Guangxia Li, Steven C. H. Hoi, Kuiyu Chang, and Ramesh Jain. Micro-blogging sentiment detection by collaborative online learning. In *ICDM*, pages 893–898, 2010.

[18] Yi Li and Philip M. Long. The relaxed online maximum margin algorithm. In *Advances in Neural Information Processing Systems (NIPS)*, pages 498–504, 1999.

[19] Nick Littlestone and Manfred K. Warmuth. The weighted majority algorithm. *Inf. Comput.*, 108(2):212–261, 1994.

[20] Francesco Orabona and Koby Crammer. New adaptive algorithms for online classification. In *NIPS*, pages 1840–1848, 2010.

[21] Frank Rosenblatt. The perceptron: A probabilistic model for information storage and organization in the brain. *Psychological Review*, 65:386–407, 1958.

[22] Volodimir G. Vovk. Aggregating strategies. In *Proceedings of the third annual workshop on Computational learning theory (COLT'90)*, pages 371–386, 1990.

[23] Jialei Wang, Peilin Zhao, and Steven C. H. Hoi. Cost-sensitive online classification. In *ICDM*, pages 1140–1145, 2012.

[24] Jialei Wang, Peilin Zhao, and Steven C. H. Hoi. Exact soft confidence-weighted learning. In *ICML*, 2012.

[25] Jialei Wang, Peilin Zhao, Steven C.H. Hoi, and Rong Jin. Online feature selection and its applications. *IEEE Transactions on Knowledge and Data Engineering*, pages 1–14, 2013.

[26] Peilin Zhao, Steven C. H. Hoi, and Rong Jin. Double updating online learning. *Journal of Machine Learning Research*, 12:1587–1615, 2011.